\documentclass{article}
\usepackage{spconf,amsmath,graphicx, amssymb}
\usepackage{booktabs}
\usepackage{CJKutf8}
\usepackage{url}
\usepackage{xcolor}
\def\isreview{0}  

\if\isreview1
    \def\papertitle{MIXTEX: DATA-EFFICIENT LATEX OCR VIA SYNTHETIC PRETRAINING AND LIMITED FINE-TUNING}
    \def\authorname{Anonymous Authors}
    \def\authoraffiliation{Anonymous Affiliation}
    
    \newcommand{\toolnamecap}{MixTeX}
    
\else
    
    \newcommand{\toolnamecap}{MixTeX}
    \def\papertitle{MIXTEX: DATA-EFFICIENT LATEX OCR VIA SYNTHETIC PRETRAINING AND LIMITED FINE-TUNING}
    \def\authorname{Yuhan Xu$^{1}$, Yijun Zhao$^{1}$, Renqing Luo$^{2}$, Gary M. Weiss$^{1}$}
\def\authoraffiliation{$^{1}$Computer and Information Sciences Department, Fordham University, New York, NY\\
$^{2}$Electrical and Computer Engineering Department, NYU Tandon School of Engineering, New York, NY}
\fi

\title{\papertitle}
\name{\authorname}
\address{\authoraffiliation}

\begin{document}
\maketitle

\begin{abstract}
LaTeX OCR converts scientific document images into editable LaTeX code. Existing systems rely on large paired datasets, which are costly to collect and limited for low-resource languages. This paper presents \toolnamecap, a data-efficient system using synthetic pretraining without real LaTeX sources. Unlike Nougat that depends on arXiv datasets, we generate training data by randomly pairing grammatical Wikipedia text with LaTeX formulas, requiring only syntactic correctness. This eliminates dependency on real document collections, enables scalable data generation (120M tokens), and supports low-resource languages. Following synthetic pretraining, adaptation requires only 400 real samples. Evaluation on a 977-sample benchmark with printed and handwritten English and Chinese shows that this two-stage strategy outperforms methods trained on large real datasets while requiring less human effort and computation. Data, code, and models are publicly available.
\end{abstract}



\section{Introduction}
\label{sec:intro}

Scientific OCR is challenging because mathematical notation is dense and sensitive to small symbol errors.
LaTeX OCR is particularly demanding as it must generate accurate, syntactically valid, LaTeX code that preserves both mathematical formulas and text formatting.
Recent end-to-end models achieve strong results when trained on large paired datasets where LaTeX sources are aligned with rendered page images, such as arXiv-derived corpora~\cite{blecher2023nougat}. However, such paired data are difficult to collect and most of the available data is in English. For Chinese documents, public LaTeX sources are scarce, making direct data collection and training challenging. This data scarcity motivates a central research question: can we build a LaTeX OCR model using synthetic pretraining that requires no real LaTeX data and only minimal real fine-tuning data?

Prior work falls into three categories. First, formula-only recognizers use attention-based models to translate cropped expressions into LaTeX~\cite{deng2017image}. Second, multi-stage pipelines integrate layout analysis, text OCR, and formula detection~\cite{li2020tablebank,smith2007overview}, though these can suffer from error propagation and higher latency~\cite{kim2022ocr}. Third, end-to-end document models employ transformer encoders with autoregressive decoders to generate markup directly from page images~\cite{blecher2023nougat,kim2022ocr}. While end-to-end models reduce error propagation, they depend heavily on large paired training corpora. This paper focuses on eliminating that data requirement through a novel synthetic pretraining strategy.
 
Synthetic data has been widely used in OCR and document understanding tasks, typically by rendering text from templates~\cite{nikolenko2021synthetic}. For LaTeX OCR specifically, existing approaches rely on rendering real LaTeX sources, such as documents collected from arXiv or academic repositories~\cite{blecher2023nougat}. Our approach differs by completely avoiding dependence on real LaTeX sources. Instead, we introduce a random pairing strategy that composes grammatical Wikipedia text with grammatical LaTeX formulas without requiring semantic alignment. This eliminates manual annotation costs and enables scalable data generation that can support training both compact models and large-scale architectures.

Unlike arXiv-derived collections, our synthetic pipeline requires no original LaTeX sources, supports multilingual Wikipedia dumps, and uses random text-formula pairing without semantic alignment. This weakens linguistic co-occurrence patterns while preserving syntactic validity, forcing reliance on visual evidence over memorized priors.

This paper describes \toolnamecap, a data-efficient LaTeX OCR system that combines synthetic pretraining with fine-tuning using limited real data. Large-scale synthetic paired data is generated by inserting grammatical LaTeX formulas into grammatical Wikipedia text, with random pairing between the two. Synthetic pretraining provides wide coverage of symbols and layouts, while a small set of 400 real examples adapts the model to actual document characteristics.

This work makes three contributions. First, it introduces a synthetic data generator that composes multilingual Wikipedia text with LaTeX formulas through random pairing to produce paired page images and LaTeX targets without manual annotation or dependence on real LaTeX sources. Second, it shows that this pretraining strategy combined with fine-tuning on limited real data achieves comparable or superior accuracy to methods trained on much larger datasets. Third, it makes publicly available a bilingual benchmark dataset of printed and handwritten content \cite{xu2026mixtexdataset}, along with code for reproducible evaluation~\cite{anonymous2024syntex}. This approach has been validated through a production deployment serving over 40,000 users.

\begin{figure}[!t]
\centering
\includegraphics[width=1\columnwidth]{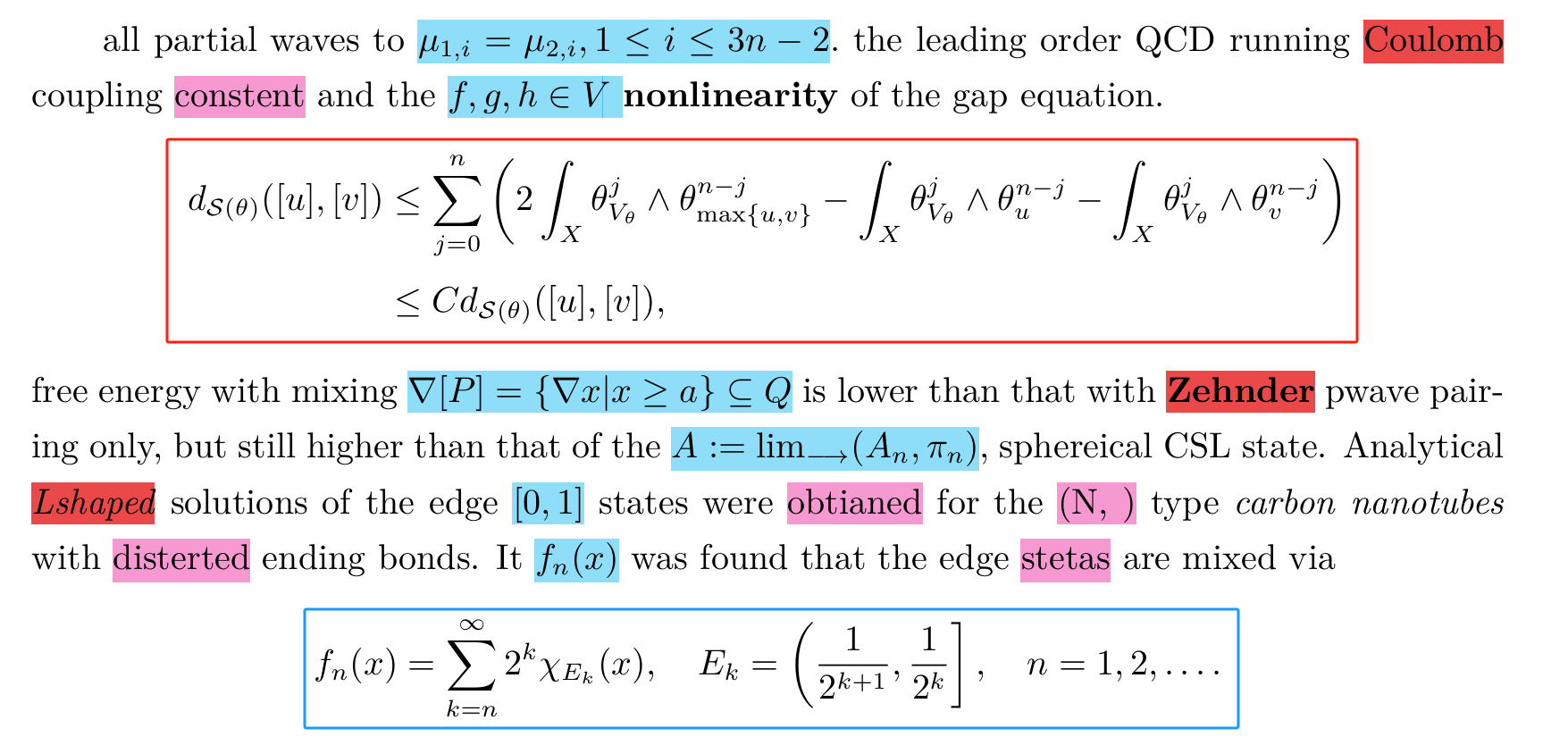}
\caption{
Training Data Sample. Original text is unhighlighted; inserted words (red), misspellings (pink), and formulas (light blue) are highlighted, with red boxes for pseudo-formulas and blue boxes for genuine expressions.
}
\vspace{-2mm}
\label{fig:data}
\end{figure}

\section{Datasets}
\label{sec:data}

\toolnamecap~utilizes three datasets: a synthetic pretraining dataset, a fine-tuning dataset, and an evaluation benchmark dataset. All three datasets include English and Chinese content. The fine-tuning and evaluation sets cover printed and handwritten modalities, whereas the synthetic set is printed-only. The synthetic dataset is generated by composing LaTeX code and compiling it to images, while the fine-tuning and evaluation datasets involve collecting real document images and manually transcribing them into verified LaTeX code.

\subsection{Synthetic Pretraining Dataset}
\label{sec:synthetic}

The goal of \toolnamecap~is to learn the mapping $f(x)$~=~$y$, for image $x$ and LaTeX code $y$ that compiles to $x$. Synthetic data generation typically involves generating instances $x$ and manually labeling them to obtain the corresponding $y$, but our approach exploits the inverse mapping property of LaTeX OCR to avoid this effort. Instead,  \toolnamecap~ synthetically generates the LaTeX code $y$ and compiles it to obtain the image $x$.


The synthetic pretraining dataset is constructed from two independent source pools: multilingual text collected from Wikipedia dumps and LaTeX formulas gathered from existing formula collections. Text segments are grammatical in the target language, and formulas are grammatical LaTeX expressions. The formula pool includes approximately 1.8M real formulas and 0.2M pseudo-formulas (see Table~\ref{tab:dataset_stats}), where pseudo-formulas are constructed by sampling valid LaTeX tokens and enforcing basic bracket balance.

To generate a synthetic page, one text excerpt is sampled and formulas are inserted at random positions. Text and formulas are randomly paired, so the page is not semantically aligned, but both parts remain syntactically valid.
To prevent the decoder from over-relying on linguistic context for next-token prediction, 
controlled corruption is applied to the sampled text, as shown in Figure~\ref{fig:data}. 
Words are randomly inserted from the same language corpus, letters within words are 
scrambled to create misspellings, and short inline formulas are inserted. These operations 
increase data diversity, prevent repetitive outputs, and force the model to rely on visual 
features rather than statistical language patterns.

\begin{table}[!t]
\centering
\caption{Synthetic Pretraining Data Statistics}
\label{tab:dataset_stats}
\small
\setlength{\tabcolsep}{4pt}
\renewcommand{\arraystretch}{0.85}
\begin{tabular}{lc}
\toprule
\textbf{Component} & \textbf{Value} \\
\midrule
Target tokens & 60M (Chinese), 60M (English) \\
Formulas (total) & 2M (1.8M real + 0.2M pseudo) \\
Rendered images & 120k \\
Avg. formulas per image & 17 \\
Avg. tokens per formula & 60 \\
Rendered Image Storage & 20GB \\
\bottomrule
\end{tabular}
\vspace{-2mm}
\end{table}

The composed LaTeX is then compiled into PDF and rasterized to images, with the LaTeX target string kept as the ground-truth sequence. Simple image normalization (crop margins, resize) is applied before training. As shown in Table~\ref{tab:dataset_stats}, the complete synthetic dataset contains approximately 120M tokens of training targets (60M Chinese, 60M English), 120k rendered page images, and 2M formulas.
On average, each image contains 17 formulas and each formula 60 tokens, with the rendered images requiring 20GB of storage.

\begin{figure}[!t]
\centering
\includegraphics[width=\columnwidth]{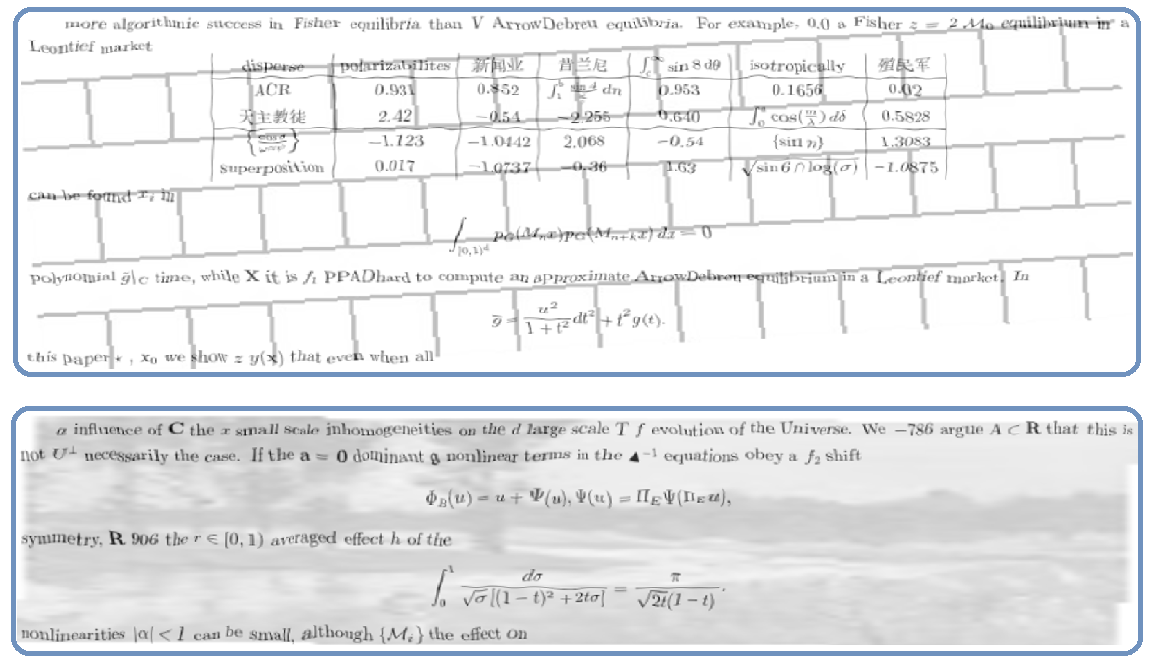}
\caption{Special cases with all types of data augmentation: noise, erosion, rotations, elastic transforms, and watermark.}
\vspace{-2mm}
\label{fig:aug}
\end{figure}

To increase robustness to style variations during pretraining, two types of augmentation are applied to the synthetic dataset. First, PDF pages are generated by compiling LaTeX with custom fonts, including handwriting-style fonts. This changes glyph appearance while keeping the LaTeX target exact. Second, image-level transformations are applied with probability 0.5 each: random Gaussian noise ($\sigma \in [5, 15]$), contrast adjustment (factor $\in [0.8, 1.2]$), morphological erosion (kernel $3 \times 3$), small rotations ($\pm 3^\circ$), elastic transforms~\cite{simard2003best} ($\alpha=30$, $\sigma=5$), and watermark overlays (opacity $\in [0.1, 0.3]$), as shown in Figure~\ref{fig:aug}. These transforms are applied only to rendered images, not target strings.

\subsection{Fine-Tuning and Evaluation Datasets}
\label{sec:real_data}

The fine-tuning and evaluation datasets were constructed using the same process and then partitioned into two disjoint sets. Unlike the synthetic approach where LaTeX is compiled to generate images, these datasets begin with real document images that are manually transcribed into syntactically valid LaTeX code.

Document images were collected from two sources: printed samples obtained by scanning mathematics and physics textbooks from university libraries, capturing diverse typesetting styles and formula densities; and handwritten samples collected from online sources where users shared mathematical notes and problem solutions. For each image, annotators manually transcribed the text and formulas into LaTeX code. The generated LaTeX was compiled to PDF and visually compared against the source image to ensure faithful reproduction of content and layout. Each sample was inspected to verify proper LaTeX syntax and mathematical correctness.

The combined fine-tuning and evaluation dataset is comprised of 1,377 samples that required 100 hours of manual annotation effort. This data set was partitioned as follows: 400 samples for fine-tuning (200 Chinese: 150 printed, 50 handwritten; 200 English: 150 printed, 50 handwritten) and 977 samples for evaluation (529 Chinese: 461 printed, 68 handwritten; 448 English: 367 printed, 81 handwritten).


\section{Experiment Methodology}
\label{sec:exp}

This section presents the experimental framework for evaluating \toolnamecap. The model architecture employs a Swin Transformer encoder with a GPT2 decoder to map document images to LaTeX sequences. Training proceeds in two stages: synthetic pretraining on 120k generated pages, followed by optional fine-tuning on 400 real samples. Performance is assessed using character-level, semantic, and token-level metrics, with comparisons against two state-of-the-art baselines.

\subsection{\toolnamecap~Model Architecture}
\label{sec:model}

\begin{figure}[!t]
\centering
\includegraphics[width=\columnwidth]{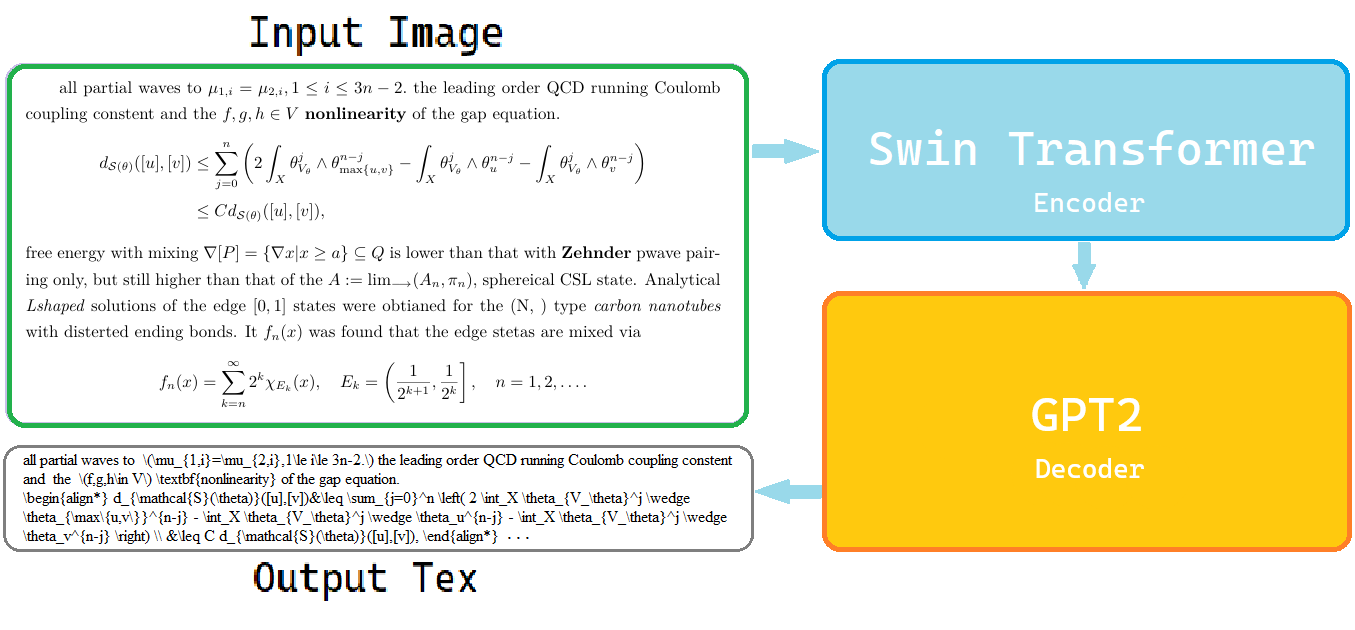}
\caption{\toolnamecap~Overview. A Swin encoder maps the image to embeddings and a GPT2 decoder generates the LaTeX code.}
\label{fig:model}
\vspace{-2mm}
\end{figure}

Figure~\ref{fig:model} shows the \toolnamecap~architecture: a Swin Transformer encoder combined with a GPT2 decoder maps document images to LaTeX sequences.
The Swin Transformer encoder scales efficiently to high-resolution images~\cite{liu2021swin}, while the GPT2 decoder handles sequential LaTeX generation~\cite{radford2019gpt2}.

The input image is resized to a fixed size and normalized. A Swin-T backbone (patch size 4, window size 7) initialized from standard pretrained weights processes the image and outputs a sequence of embeddings $\{\mathbf{z}_i\}_{i=1}^{n}$. The decoder is a compact GPT2 architecture with hidden size 768, 12 attention heads, and 4 transformer layers. It attends to encoder embeddings through cross-attention and generates tokens autoregressively. A Byte Pair Encoding (BPE) tokenizer is trained on the training corpus to cover English, Chinese, and LaTeX commands, and is used during evaluation.

\subsection{Training Setup}
\label{sec:train}

Two model variants are generated: \toolnamecap-Pretrain, which is trained only on the synthetic dataset, and \toolnamecap-Finetune, where synthetic pretraining is followed by fine-tuning. As described in Section~\ref{sec:real_data}, the fine-tuning dataset contains 400 English and Chinese samples, covering printed and handwritten content.
Input images are resized to $500 \times 400$ with a maximum output length of 296 tokens. Pretraining is run to convergence with batch size 24 on a single RTX 4090. AdamW optimizer is used with mixed precision, learning rate $1.3 \times 10^{-4}$ decayed to $7 \times 10^{-6}$ via cosine annealing. Fine-tuning uses the same optimizer with initial learning rate $5 \times 10^{-5}$ and the same decay schedule.

\begin{table*}[t]
\centering
\footnotesize
\caption{Quantitative Results on Printed and Handwritten Data}
\label{tab:results}
\setlength{\tabcolsep}{3pt}
\begin{tabular}{lcccc@{\hspace{1.5em}}cccc}
\toprule
\multicolumn{5}{c}{\textbf{English (Printed)}} & \multicolumn{4}{c}{\textbf{English (Handwritten)}} \\
\cmidrule(r){1-5} \cmidrule(l){6-9}
Model & NED$\downarrow$ & METEOR$\uparrow$ & F1$\uparrow$ & Precision$\uparrow$ & NED$\downarrow$ & METEOR$\uparrow$ & F1$\uparrow$ & Precision$\uparrow$ \\
\midrule
\toolnamecap-Pretrain~~~ & 0.339$\pm$0.206 & 0.357$\pm$0.303 & 0.820$\pm$0.148 & 0.812$\pm$0.186 & 0.829$\pm$0.143 & 0.059$\pm$0.115 & 0.346$\pm$0.256 & 0.428$\pm$0.306 \\
\toolnamecap-Finetune~~~ & \textbf{0.036$\pm$0.051} & \textbf{0.927$\pm$0.140} & \textbf{0.981$\pm$0.028} & \textbf{0.985$\pm$0.028} & \textbf{0.169$\pm$0.262} & \textbf{0.800$\pm$0.265} & \textbf{0.878$\pm$0.257} & \textbf{0.897$\pm$0.262} \\
Nougat~~~ & 0.635$\pm$0.413 & 0.359$\pm$0.297 & 0.757$\pm$0.294 & 0.827$\pm$0.295 & 0.913$\pm$0.185 & 0.176$\pm$0.256 & 0.303$\pm$0.332 & 0.534$\pm$0.430 \\
Hunyuan~~~ & 0.347$\pm$0.159 & 0.335$\pm$0.333 & 0.829$\pm$0.106 & 0.896$\pm$0.122 & 0.528$\pm$0.237 & 0.368$\pm$0.259 & 0.623$\pm$0.267 & 0.783$\pm$0.313 \\
\midrule
\multicolumn{5}{c}{\textbf{Chinese (Printed)}} & \multicolumn{4}{c}{\textbf{Chinese (Handwritten)}} \\
\cmidrule(r){1-5} \cmidrule(l){6-9}
Model & NED$\downarrow$ & METEOR$\uparrow$ & F1$\uparrow$ & Precision$\uparrow$ & NED$\downarrow$ & METEOR$\uparrow$ & F1$\uparrow$ & Precision$\uparrow$ \\
\midrule
\toolnamecap-Pretrain~~~ & 0.431$\pm$0.219 & 0.257$\pm$0.209 & 0.767$\pm$0.187 & 0.766$\pm$0.211 & 0.940$\pm$0.082 & 0.007$\pm$0.031 & 0.129$\pm$0.176 & 0.168$\pm$0.239 \\
\toolnamecap-Finetune~~~ & \textbf{0.064$\pm$0.070} & \textbf{0.875$\pm$0.130} & \textbf{0.966$\pm$0.036} & \textbf{0.975$\pm$0.042} & \textbf{0.528$\pm$0.384} & \textbf{0.346$\pm$0.313} & \textbf{0.539$\pm$0.421} & 0.536$\pm$0.426 \\
Nougat~~~ & 0.874$\pm$0.258 & 0.029$\pm$0.052 & 0.442$\pm$0.290 & 0.719$\pm$0.351 & 0.968$\pm$0.057 & 0.002$\pm$0.002 & 0.015$\pm$0.045 & 0.064$\pm$0.193 \\
Hunyuan~~~ & 0.404$\pm$0.119 & 0.193$\pm$0.175 & 0.809$\pm$0.098 & 0.881$\pm$0.112 & 0.742$\pm$0.236 & 0.037$\pm$0.084 & 0.367$\pm$0.311 & \textbf{0.550$\pm$0.416} \\
\bottomrule
\end{tabular}
\end{table*}

\subsection{Metrics}
\label{sec:metrics}

Four metrics are used to evaluate model performance: Normalized Edit Distance (NED), METEOR, and token-level precision and F1 score. To illustrate these metrics, a very simple example is used where the ground truth is ``\verb|\frac{1}{2}|'' but the model predicts ``\verb|\frac{1}{3}|''.

Normalized Edit Distance  measures character-level accuracy by computing the edit distance between predicted and ground-truth LaTeX strings, normalized by the length of the longer sequence. Let $s$ be the predicted LaTeX string, $t$ the ground-truth string, and $\mathrm{ED}(s,t)$ the character-level Levenshtein distance ~\cite{levenshtein1966binary}  computed at the Unicode character level by treating LaTeX commands as sequences of characters. Then Normalized Edit distance is defined by Equation~\ref{eq:ned}. In the simple example, $\mathrm{ED}$ = 1 (one character differs: ``2'' vs ``3'') and $\mathrm{NED}$ = 1/11 $\approx$ 0.091. Lower NED values indicate better performance ($\downarrow$). 
\begin{equation}
\label{eq:ned}
\mathrm{NED}(s,t) = \frac{\mathrm{ED}(s,t)}{\max(|s|,|t|)}
\end{equation}

METEOR~\cite{banerjee2005meteor} evaluates semantic alignment quality between predicted and ground-truth sequences by considering exact token matches, synonym matching, and word ordering. This captures semantic correctness beyond character-level accuracy. METEOR is computed using the model's BPE tokenizer for both English and Chinese, with stemming and synonym matching disabled to preserve LaTeX command integrity. Higher values indicate better performance ($\uparrow$).

Token-level metrics measure overlap between prediction and ground truth at the tokenized level. Both prediction $s$ and ground truth $t$ are tokenized using the model's BPE tokenizer, yielding token multisets $S = \{s_1, \ldots, s_m\}$ and $T = \{t_1, \ldots, t_n\}$. Token overlap is computed treating sequences as multisets (position-invariant). The number of True Positives is defined by Equation~\ref{eq:tp},
where $\text{count}_S(\text{tok})$ and $\text{count}_T(\text{tok})$ denote the occurrence frequency of token $\text{tok}$ in sequences $S$ and $T$, respectively. 
\begin{equation}
\label{eq:tp}
\text{TP} = \sum_{\text{tok}} \min(\text{count}_S(\text{tok}), \text{count}_T(\text{tok}))
\end{equation}
The false positives are computed as $\text{FP} = |S| - \text{TP}$ and the false negatives as $\text{FN} = |T| - \text{TP}$. Precision, recall, and F1 are computed from these in the standard way~\cite{powers2011evaluation}. In the example, when tokenized, 
$T$ = \{\verb|\frac|, \verb|{|, \verb|1|, \verb|}|, \verb|{|, \verb|2|, \verb|}|\} and $S$ = \{\verb|\frac|, \verb|{|, \verb|1|, \verb|}|, \verb|{|, \verb|3|, \verb|}|\}. 
Also, $|S| = |T| = 7$ and 6 of the 7 tokens match, giving $\text{TP}$ = 6, $\text{FP}$ = 1, $\text{FN}$ = 1, which yields a precision, recall, and F1 of 6/7. Position-invariant matching may overestimate accuracy for structure-sensitive LaTeX. Higher values indicate better performance ($\uparrow$).

\subsection{Baseline Methods}
\label{sec:baseline}

Two state-of-the-art LaTeX OCR systems are used as baselines: Nougat~\cite{blecher2023nougat} and Hunyuan OCR~\cite{hunyuanvisionteam2025hunyuanocrtechnicalreport}. Hunyuan is a commercial-grade OCR system with strong multilingual support. For Nougat, the official nougat-base checkpoint and inference pipeline are used without modifications, keeping all default settings. For Hunyuan, the open-source Hunyuan OCR system is used with default parameters. We compare models under representative training paradigms: arXiv-based training for Nougat versus synthetic pretraining with limited real fine-tuning for \toolnamecap.
Repository~\cite{anonymous2024syntex} commit hashes and checkpoint identifiers are recorded for exact reproducibility.

\vspace{-.1in}
\section{Results}
\label{sec:results}
This section provides a  quantitative and qualitative comparison of the results for \toolnamecap-Pretrain and \toolnamecap-Finetune versus the two baseline methods.

\begin{figure*}[h!]
\centering
\includegraphics[width=2\columnwidth]{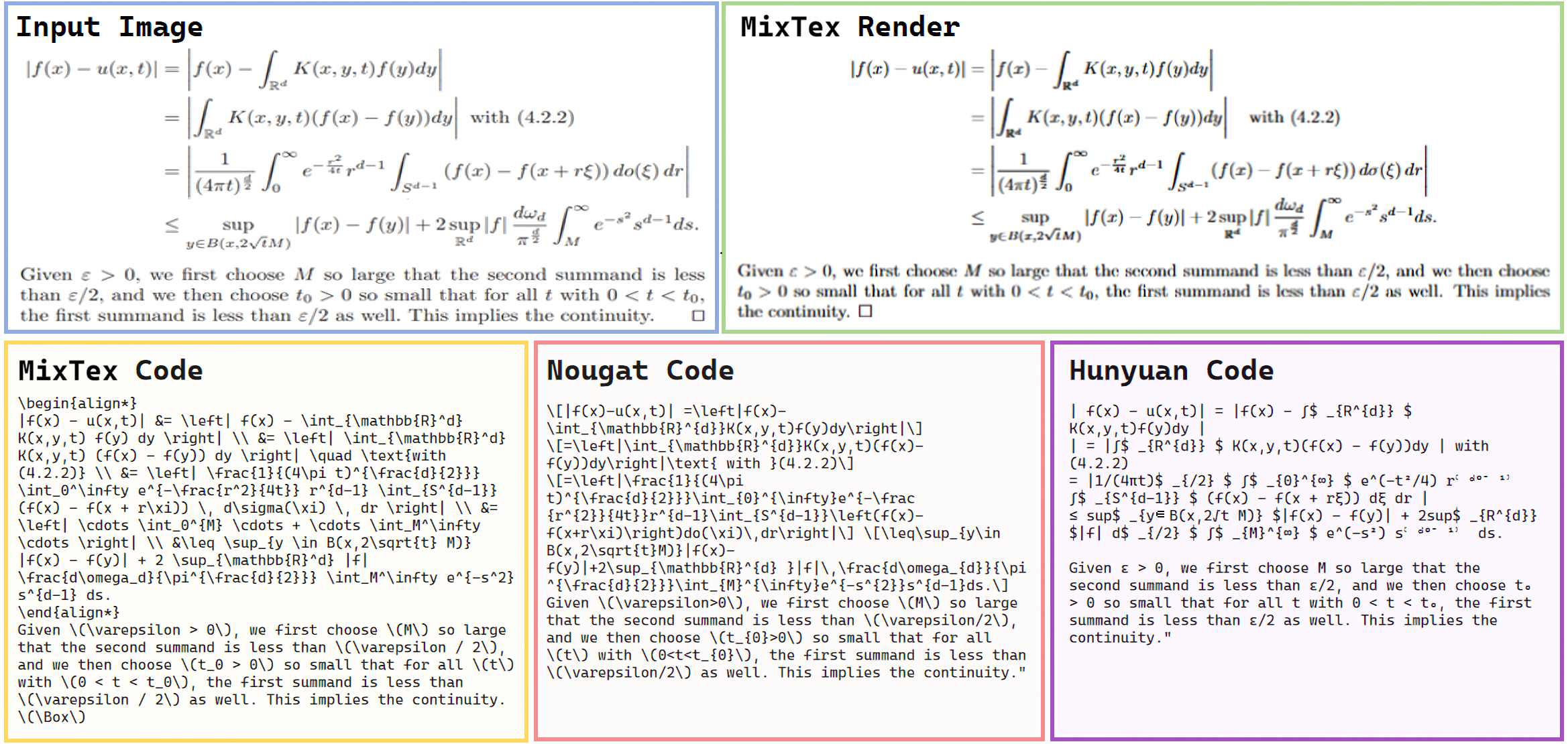}
\vspace{-2mm}
\caption{Qualitative comparison on a mathematical proof. Top: input image (left) and our rendered output (right). Bottom: LaTeX code generated by \toolnamecap~(left), Nougat (middle), and Hunyuan (right).
}
\label{fig:qualitative}
\end{figure*}

\subsection{Quantitative Results}
\label{sec:quant}

The main quantitative results for \toolnamecap~pretrained and fine-tuned models, along with Nougat and Hunyuan baselines, are in Table~\ref{tab:results}. The values are mean $\pm$ standard deviation, with the best model performance highlighted in bold.

Several observations emerge. \toolnamecap-Finetune achieves best performance in every case except one (Chinese handwritten precision), where it is a close second, while in most cases it substantially outperforms all other methods. Based on these large performance differences and reported standard deviations, it is clear that \toolnamecap-Finetune outperforms other methods with statistical confidence (t-test results omitted  due to space constraints). \toolnamecap-Pretrain shows mixed results: on printed content it is competitive with or outperforms Nougat but underperforms Hunyuan; on handwritten content it performs poorly, demonstrating need for real fine-tuning data. For printed English, \toolnamecap-Finetune substantially outperforms Nougat and Hunyuan on all metrics (e.g., NED of 0.036 vs. 0.635 and 0.347). For printed Chinese the improvement is more pronounced, demonstrating the benefit of multilingual synthetic pretraining for low-resource languages.

On handwritten data, the task is significantly harder, but \toolnamecap-Finetune still yields the best results. For English handwriting, \toolnamecap-Finetune achieves NED of 0.169 while \toolnamecap-Pretrain struggles with NED of 0.829, showing that 50 handwritten fine-tuning samples yields dramatic improvement. Chinese handwriting remains most challenging, with \toolnamecap-Finetune achieving NED of only 0.528, but still substantially outperforms both baselines.

\subsection{Qualitative Comparison}
\label{sec:qual}

Figure~\ref{fig:qualitative} compares all three methods on a mathematical proof with integrals, fractions, Greek symbols, and multi-line equations. This example illustrates a challenging case; generalization is primarily validated by the 977-sample quantitative benchmark. \toolnamecap-Finetune correctly recognizes the full structure, producing LaTeX that renders identically (top right) to the input, with proper handling of nested structures like \texttt{$\backslash$frac}, alignment operators, and mathematical symbols.
In contrast, Nougat's code contains numerous errors: incorrect spacing, character recognition mistakes, and malformed commands (e.g., ``\texttt{$\backslash$left[}'' without matching delimiter). Hunyuan produces syntactically valid LaTeX but misses critical structural details, such as integral bounds and equation references. Hunyuan sometimes mixes Unicode symbols with LaTeX commands; character-to-LaTeX substitutions are applied during evaluation, improving its reported metrics.

The synthetic pipeline also aids downstream editing: controlled pretraining conventions produce stylistically consistent outputs, reducing post-OCR correction burden, particularly for multilingual documents.




\section{Conclusion}
\label{sec:conclusion}

This paper presents \toolnamecap{}, a data-efficient LaTeX OCR system with a synthetic pretraining approach requiring no real LaTeX sources. Unlike Nougat, which relies on large-scale real paired datasets from arXiv-derived sources, \toolnamecap{} generates training data by randomly pairing grammatical Wikipedia text with grammatical LaTeX formulas, without semantic alignment. This strategy enforces syntactic correctness, removes the need for curated LaTeX document collections, supports scalable data generation for compact and large architectures, and improves accessibility for low-resource languages. After synthetic pretraining, fine-tuning on only 400 real samples adapts the model to real document characteristics.

Evaluation on a bilingual benchmark of 977 annotated samples shows that \toolnamecap{} outperforms Nougat and Hunyuan across nearly all scenarios, despite those baselines using much larger real paired datasets. This data efficiency makes the pipeline accessible to individual researchers and offers a replicable methodology for document understanding tasks where paired data is expensive or scarce.

A key limitation is that small syntactic errors, such as a missing bracket, can cause LaTeX formulas to fail rendering, while current token-level metrics inadequately penalize such critical errors. Future work will incorporate reinforcement learning with compilation-based rewards to optimize syntactic correctness. Extending synthetic generation to full document structures, including multi-column layouts, citations, tables, and algorithms, is a promising direction for full-document LaTeX OCR.

\bibliographystyle{IEEEbib}
\bibliography{refs}

@article{blecher2023nougat,
  title={Nougat: Neural Optical Understanding for Academic Documents},
  author={Blecher, Lukas and others},
  journal={arXiv preprint arXiv:2308.13418},
  year={2023}
}

@inproceedings{deng2017image,
  title={Image-to-Markup Generation with Coarse-to-Fine Attention},
  author={Deng, Yuntian and Kanervisto, Anssi and Ling, Jeffrey and Rush, Alexander M.},
  booktitle={Proceedings of ICML},
  year={2017}
}

@inproceedings{kim2022ocr,
  title={OCR-free Document Understanding Transformer},
  author={Kim, Geewook and others},
  booktitle={Proceedings of ECCV},
  year={2022}
}

@misc{anonymous2024syntex,
  title = {MixTeX code and data repository},
author = {Xu, Yuhan and Zhao, Yijun and Luo, Renqing and Weiss, Gary M.},
year = {2025},
  howpublished = {\url{https://github.com/yuhanxu01/MixTeX}}
}

@misc{xu2026mixtexdataset,
  title     = {MixTeX Synthetic Pretraining Dataset},
  author    = {Xu, Yuhan and Zhao, Yijun and Luo, Renqing and Weiss, Gary M.},
  year      = {2026},
  month     = may,
  publisher = {Zenodo},
  howpublished       = {\url{https://doi.org/10.5281/zenodo.20017670}}
}

@inproceedings{liu2021swin,
  title={Swin Transformer: Hierarchical Vision Transformer using Shifted Windows},
  author={Liu, Ze and Lin, Yutong and Cao, Yue and Hu, Han and Wei, Yixuan and Zhang, Zheng and Lin, Stephen and Guo, Baining},
  booktitle={Proceedings of ICCV},
  year={2021}
}

@inproceedings{li2020tablebank,
  title={TableBank: Table Benchmark for Image-based Table Detection and Recognition},
  author={Li, Minghao and others},
  booktitle={Proceedings of LREC},
  year={2020}
}

@inproceedings{smith2007overview,
  title={An Overview of the Tesseract OCR Engine},
  author={Smith, Ray},
  booktitle={Proceedings of ICDAR},
  year={2007}
}

@book{nikolenko2021synthetic,
  title={Synthetic Data for Deep Learning},
  author={Nikolenko, Sergey},
  publisher={Springer},
  year={2021}
}

@article{radford2019gpt2,
  title={Language Models are Unsupervised Multitask Learners},
  author={Radford, Alec and Wu, Jeffrey and Child, Rewon and Luan, David and Amodei, Dario and Sutskever, Ilya},
  journal={OpenAI Technical Report},
  year={2019}
}

@article{levenshtein1966binary,
  title={Binary Codes Capable of Correcting Deletions, Insertions, and Reversals},
  author={Levenshtein, Vladimir I.},
  journal={Soviet Physics Doklady},
  volume={10},
  number={8},
  pages={707--710},
  year={1966}
}

@inproceedings{banerjee2005meteor,
  title={METEOR: An Automatic Metric for MT Evaluation with Improved Correlation with Human Judgments},
  author={Banerjee, Satanjeev and Lavie, Alon},
  booktitle={Proceedings of the ACL Workshop on Intrinsic and Extrinsic Evaluation Measures for Machine Translation},
  year={2005}
}

@misc{hunyuanvisionteam2025hunyuanocrtechnicalreport,
      title={HunyuanOCR Technical Report}, 
      author={Hunyuan Vision Team},
      year={2025},
      eprint={2511.19575},
      archivePrefix={arXiv},
      primaryClass={cs.CV},
      url={https://arxiv.org/abs/2511.19575}, 
}

@inproceedings{simard2003best,
  title={Best practices for convolutional neural networks applied to visual document analysis},
  author={Simard, Patrice Y and Steinkraus, Dave and Platt, John C},
  booktitle={Proceedings of the Seventh International Conference on Document Analysis and Recognition (ICDAR)},
  volume={2},
  pages={958--963},
  year={2003},
  organization={IEEE}
}

@article{powers2011evaluation,
  title={Evaluation: from precision, recall and F-measure to ROC, informedness, markedness and correlation},
  author={Powers, David Martin W},
  journal={Journal of Machine Learning Technologies},
  volume={2},
  number={1},
  pages={37--63},
  year={2011}
}

\end{document}